\PassOptionsToPackage{table}{xcolor} 

\documentclass[preprint,12pt, a4paper]{elsarticle}

\usepackage{amssymb}
\usepackage{hyperref}
\usepackage{orcidlink} 
\usepackage{booktabs}
\usepackage{tabularx}
\usepackage{array, makecell, booktabs}
\usepackage{makecell}   
\usepackage{nomencl}
\makenomenclature
\usepackage{framed} 
\usepackage{subcaption}
\usepackage{appendix}

\journal{SoftwareX}

\begin{document}
\renewcommand{\labelenumii}{\arabic{enumi}.\arabic{enumii}}

\begin{frontmatter}
\title{AI4EF: Artificial Intelligence for Energy Efficiency in the Building Sector}



\author{Alexandros Menelaos Tzortzis\orcidlink{0009-0002-5020-4485}}
\author{Georgios Kormpakis\orcidlink{0000-0003-4052-4549}}
\author{Sotiris Pelekis\orcidlink{0000-0002-9259-9115}}
\author{Ariadni Michalitsi-Psarrou\orcidlink{0000-0003-0792-8568}}
\author{Evangelos Karakolis\orcidlink{0000-0003-2833-3088}}
\author{Christos Ntanos\orcidlink{0000-0002-5162-6500}}
\author{Dimitris Askounis\orcidlink{0000-0002-2618-5715}}

\address{Decision Support Systems Laboratory\\
School of Electrical and Computer Engineering\\
Institute of Communications and Computer Systems\\
National Technical University of Athens\\
Greece}


\begin{abstract}

AI4EF (Artificial Intelligence for Energy Efficiency) is an advanced, user-centric tool designed to support decision-making in building energy retrofitting and efficiency optimization. Leveraging machine learning (ML) and data-driven insights, AI4EF (Artificial Intelligence for Energy Efficiency) enables stakeholders—such as public sector representatives, energy consultants, and building owners—to model, analyse, and predict energy consumption, retrofit costs, and environmental impacts of building upgrades. Featuring a modular framework, AI4EF includes customizable building retrofitting, photovoltaic installation assessment, and predictive modelling tools that allow users to input building parameters and receive tailored recommendations for achieving energy savings and carbon reduction goals. Additionally, the platform incorporates a Training Playground for data scientists to refine ML models used by said framework. Finally, AI4EF provides access to the Enershare Data Space to facilitate seamless data sharing and access within the ecosystem. AI4EF’s compatibility with open-source identity management (Keycloak) enhances security and accessibility, making it adaptable for various regulatory and organizational contexts. This paper presents an overview of AI4EF’s architecture, its application in energy efficiency scenarios, and its potential for advancing sustainable energy practices through artificial intelligence (AI).
\end{abstract}


\end{frontmatter}


\section*{Metadata}
\label{metadata}


\begin{table}[!h]
    \centering
    \caption{Code and Software Metadata}
    \footnotesize 
    \begin{tabularx}{\linewidth}{@{}lX@{}} 
        \toprule \\
        \textbf{Code Metadata} & \\
        \hline
        Current code version & 0.0.2 \\
        Permanent link to code/repository & \url{https://github.com/epu-ntua/enershare-ai4ef} \cite{zenodo-enershare-ai4ef} \\
        Code Ocean compute capsule & -- \\
        Legal Code License & Attribution-Non-commercial 4.0 International \\
        Code versioning system used & Git \cite{git}\\
        Software code languages, tools, and services used & Python \cite{python}, JavaScript, Docker \\
        Compilation requirements, operating environments \& dependencies & Docker \\
        Link to developer documentation/manual & \url{https://github.com/epu-ntua/enershare-ai4ef/blob/master/README.md} \\ 
        API documentation & \url{https://epu-ntua.github.io/enershare-ai4ef/} \\ 
        Support email for questions & \url{atzortzis@epu.ntua.gr} \\
        \hline \\
        \textbf{Software Metadata} & \\
        \hline
        Current software version & 0.0.2 \\
        Permanent link to executables of this version & \url{https://github.com/epu-ntua/enershare-ai4ef/releases/tag/0.0.2} \\
        Legal Software License & Attribution-Non-commercial 4.0 International \\
        Computing platforms/Operating Systems & Linux, Web-based \\
        Installation requirements \& dependencies & Docker for Keycloak server, pip \cite{pypi} for AI4EF frontend and backend \\
        Link to user manual & \url{https://github.com/epu-ntua/enershare-ai4ef/blob/master/README.md} \\
        Support email for questions & \url{atzortzis@epu.ntua.gr} \\
        \bottomrule
    \end{tabularx}
    \label{tab:metadata}
\end{table}

\begin{framed}
    \printnomenclature
\end{framed}

\nomenclature{ML}{Machine Learning}
\nomenclature{HPO}{Hyper Parameter Optimization}
\nomenclature{MLP}{Multi-layer perceptron}
\nomenclature{IDS}{International Data Spaces}
\nomenclature{HVAC}{Heating, Ventilation, and Air Conditioning}
\nomenclature{MLOps}{Machine Learning Operations}
\nomenclature{GHG}{Green House Gas}
\nomenclature{SPA}{Single Page Application}

\section{Motivation and significance}
The European Union (EU) is committed to combating climate change and meeting the 2015 Sustainable Development Goals (SDGs) \cite{EU_SDGs}\cite{EU_EED_Modelling}. Energy efficiency is a key priority, with a target to reduce energy consumption by 13\% by 2030 \cite{EU_EED_Modelling}. The REPowerEU plan \cite{eu_repowereu_2024} focuses on increasing building renovation rates, especially deep renovations, and accelerating heat pump deployment to replace fossil fuel boilers, thereby reducing fossil fuel dependence and bolstering energy security.
In line with these goals, the EU emphasizes renewable energy in sustainable building design. Technologies like solar panels, wind turbines, and biomass systems decrease reliance on fossil fuels and support a sustainable energy future \cite{Shi2012}\cite{doi:10.1177/0270467602022002003}\cite{su14084792}\cite{Kalair2020}. Given that buildings contribute roughly 40\% of global $CO_2$ emissions, this shift is essential for sustainable urban design \cite{Zhang2024}.

Moreover, building retrofitting is crucial for sustainable architecture, as outdated structures contribute significantly to energy consumption, with lighting, heating, cooling, and air conditioning systems accounting for 16-50\% of global energy use \cite{Pombo2016}. Many European buildings, especially those from the 1940s-1970s, have poor energy performance due to inadequate insulation and outdated heating systems \cite{Tarja2012}. Retrofitting these buildings by upgrading insulation, installing energy-efficient windows \cite{ijerph19021016}, green roofs \cite{ziaee2022}, and using smart materials \cite{RAGHEB2016778}\cite{DODOO20111589}\cite{ObinnaIwuanyanwu2024}\cite{MOHAMED2017139}\cite{PATIL20221813}, alongside replacing old heating systems \cite{TosinMichaelOlatunde2024}\cite{pramanik2020green}, can greatly enhance energy efficiency, reduce GHG emissions, and improve indoor air quality. Moreover, real-time energy monitoring with smart technologies can optimize consumption \cite{Tarja2012}\cite{Hilger2018}\cite{pramanik2020green}. With supportive policies and incentives, retrofitting offers a significant opportunity to reduce environmental impact, improve comfort, and generate long-term savings.

Building on the EU’s energy goals and the shift toward renewable and energy-efficient building solutions, artificial intelligence (AI) plays an increasingly pivotal role in optimizing these sustainable energy systems. As the world transitions from fossil fuels to renewable energy sources (RES) like solar, wind, and geothermal energy \cite{RAHMAN2022112279}, AI enhances efficiency, reduces emissions, and mitigates climate impacts \cite{Ukoba2024}\cite{ZHOU2024107355}\cite{DELANOE2023117261}. ML algorithms optimize renewable energy infrastructure by analyzing data such as smart meter readings and weather patterns to improve operations and reduce downtime \cite{Toscano2024}\cite{Ukoba2024}. AI also supports infrastructure inspections, maintenance, and climate risk assessments, enhancing system reliability \cite{Farzaneh2021}\cite{Irfan2024}\cite{MEHMOOD2019109383}. Despite challenges like limited investment, slow permitting, supply chain issues, and data privacy concerns \cite{8481346}\cite{MASSOUDAMIN2011547}, a strategic AI-driven approach combined with supportive policies is vital for achieving net-zero emissions by 2050 \cite{DELANOE2023117261}\cite{JRC118592}.

This paper presents AI4EF, a software platform designed to support decision-making in energy efficiency. Its backend consists of modular services tailored to specific user needs, such as investment planning and photovoltaic installation, providing targeted functionalities. For users requiring custom models, AI4EF offers a separate service for creating and executing ML pipelines, allowing the development and evaluation of models with personalized datasets. By applying advanced machine learning operations' (MLOps) practices, this service simplifies the model development and deployment process, making it accessible to both data scientists and energy consultants, and empowering informed decisions in energy efficiency.

AI4EF addresses key limitations of both open-source and proprietary tools in the building and renewable energy sectors. The primary contributions of this tool are as follows:
\begin{itemize}
    \item \textbf{Open-Source Accessibility:} This transparency enables users to understand, customize, and enhance its functionality. In contrast to many proprietary tools \cite{equest,buildingiq,revit_dynamo,aurora_solar} that restrict user customization, our tool provides the flexibility to tailor it to individual requirements.
    \item \textbf{Customization through AI Model Training:} The tool enables users to train custom AI models on their own datasets, offering highly personalized estimations and recommendations. This user-driven approach contrasts with other open-source tools, such as OpenStudio\cite{openstudio}, which primarily focuses on predefined energy modelling without extensive support for user-defined model training. Additionally, proprietary tools like Retrofit Advisor\cite{retrofit_advisor} and HelioScope\cite{helioscope} are restricted to fixed, vendor-controlled algorithms, limiting their flexibility for users seeking customized solutions. 
    \item \textbf{Real-World Validation and User-Centric Design:} This tool was built on real data based on the needs of Latvian Environmental Investment Fund (LEIF) \cite{LEIF}, and was evaluated and utilised by real users satisfying their needs. This ensures that the tool meets real-world demands and offers a practical solution for issues regarding building energy efficiency.
    \item \textbf{Integration with the European Energy Data Space:} AI4EF is built on the Enershare Data Space \cite{enershare}, a core element of the European energy data ecosystem. By utilizing a verified IDS (International Data Spaces) connector, AI4EF enables collaborative innovation through access to diverse, large-scale datasets. In contrast, tools like Solar-Log \cite{solarlog} and Aurora Solar \cite{aurora_solar} rely on limited datasets, restricting analysis depth.
    \item \textbf{User-Friendly Insights with Minimal Data:} AI4EF provides quick retrofit and PV recommendations using minimal building data, making it accessible to users with limited information. Unlike other retrofitting tools \cite{openstudio,equest,buildingiq,retrofit_advisor,revit_dynamo}, which require extensive input on building geometry, Heating, Ventilation, and Air Conditioning (HVAC), and other details, AI4EF offers initial insights with only a few key features. Similarly, PV-focused tools \cite{pvsol,helioscope,aurora_solar} require comprehensive data for detailed simulations, making them less accessible for users lacking extensive building data.
\end{itemize}

\section{Software description}


AI4EF is an AI-based tool that combines robust data analysis, ML, and accessible modelling interfaces, to offer actionable insights into renewable energy integration and building retrofitting. The software architecture, depicted in figure \ref{fig:architecture}, illustrates the interactions between the frontend interface, analytical modules, and ML back-end, highlighting how each component contributes to a seamless, integrated user experience.

\subsection{Software architecture}

The AI4EF platform is designed using Docker \cite{docker}, with a modular micro-services architecture to enhance scalability, maintainability, and functionality. It comprises of four main components, each fulfilling distinct roles within the platform

\paragraph{\textbf{Identity and access management} (Optional)} AI4EF uses Keycloak \cite{keycloak} for secure user authentication and centralized identity management. With Keycloak, the platform supports single sign-on (SSO) based on the OpenID Connect (OIDC) protocol for seamless access to services and role-based permissions to ensure that only authorized users can access specific data and features.

\paragraph{\textbf{Frontend Application}}


The AI4EF dashboard is a Single Page Application (SPA) developed with React \cite{react}, providing a modern, efficient approach to web development. Unlike traditional Multi-Page Applications, it loads essential resources (HTML\cite{html}, CSS\cite{css}, and JavaScript\cite{javascript}) once, enabling faster performance and dynamic content updates. This design improves caching, simplifies debugging, and enhances user experience. The dashboard is responsive, ensuring seamless use across desktop and mobile devices. It leverages the MUI library \cite{mui_library} for cohesive styling and UI components, and integrates with Keycloak \cite{keycloak} for secure user authentication and streamlined access to AI4EF’s analytical tools.

\paragraph{\textbf{MLApp (Back-end)}}
The backend component serves as the core foundation of the AI4EF platform, It contains distinct, independently deployed sub-services for \textit{Building Retrofitting} and \textit{Photovoltaic (PV) Installation} analysis. It's responsible for (a) retrieving data from the internal filesystem, (b) preprocessing and harmonizing this data to meet the input requirements of the forecasting model, (c) managing the storage, versioning, and deployment of the model, and (d) handling user requests by providing appropriate responses. 

\paragraph{\textbf{Training Playground}}
This service establishes an MLOps environment to manage data flows and orchestrate model training and tuning pipelines. It enables users to build and customize models for building efficiency goals, ensuring seamless integration with AI4EF’s MLApp and other modules. The service handles (a) data fetching and preprocessing, (b) model training and hyperparameter optimization, (c) performance evaluation, and (d) model storage, versioning, and deployment. 

\subsubsection{Technologies}

The key technologies that power AI4EF are as follows:

\begin{itemize}

  \item \textbf{Dagster} \cite{dagster}: Provides the overall workflow and pipeline management framework, for modular, scalable, and robust data processing, model training, and deployment.
  \item \textbf{Scikit-learn} \cite{scikitLearn}: Employed in the data preprocessing phase for scaling, managing categorical variables, and handling missing values, as well as contributing to data compatibility and standardization across models.
  \item \textbf{Scipy} \cite{sciPy}: Used in data extraction, cleaning, and scaling, complementing Scikit-learn's preprocessing functionalities to structure data for model training.
  \item \textbf{PyTorch Lightning} \cite{pyTorchLightning}: The primary deep learning framework, providing a streamlined structure for defining, training, and evaluating neural network architectures used in the tool’s sub-services.
  \item \textbf{Optuna} \cite{optuna}: Used for hyperparameter optimization (HPO) in training, where it systematically tunes parameters to improve model performance. employing pruning techniques to discard suboptimal trials early.
  \item \textbf{Darts} \cite{darts}: During the evaluation phase it assess model predictions, generating metrics for regression analysis.
  \item \textbf{Data space connector} \cite{tsg_connector}: Supports data acquisition for the training playground, expanding accessible data sources tested using the Enershare Data Space .   
  \item \textbf{FastAPI} \cite{fastAPI}: Deploys API endpoints for model and UI integration, processing incoming data requests from the front-end, wrangling the data, and returning model predictions.
\end{itemize}

\subsubsection{European Energy Data Space Integration} 

\begin{figure}[!h]
    \centering
    \includegraphics[width=\linewidth]{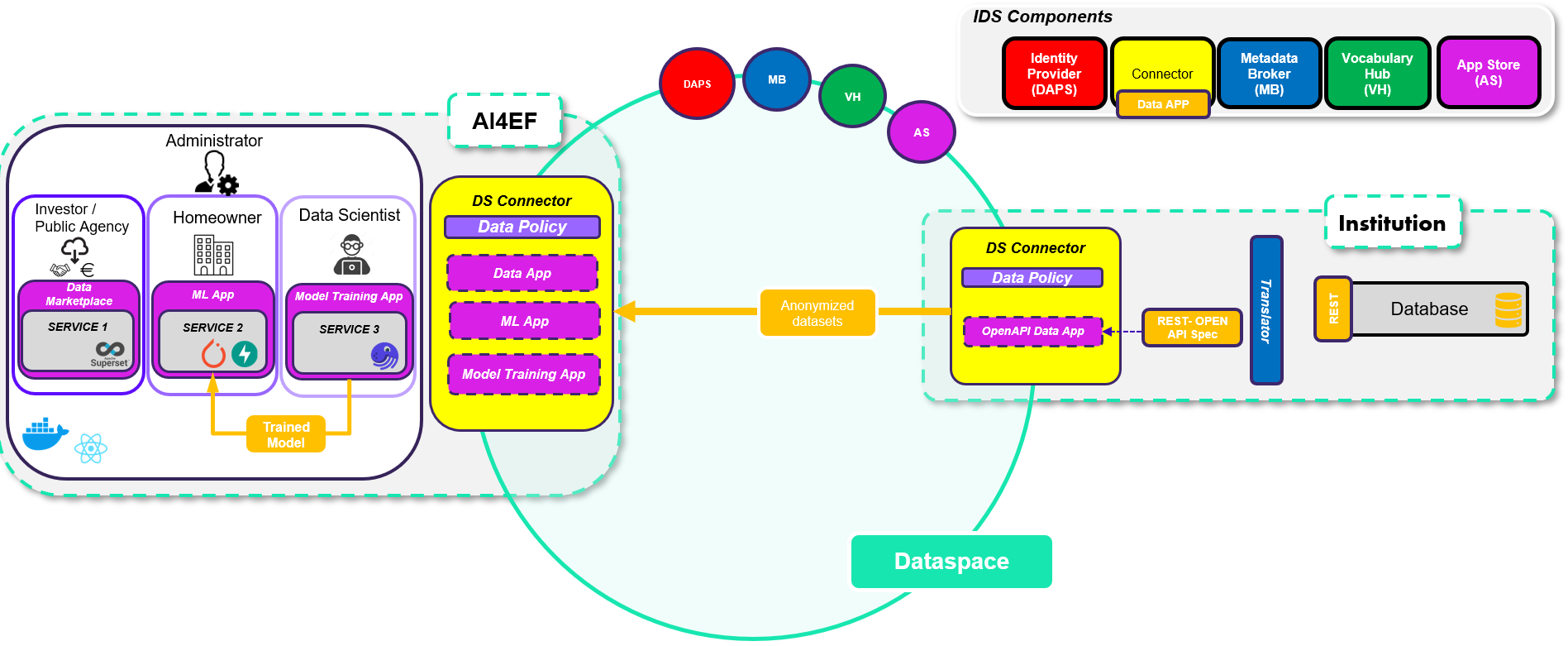}
    \caption{AI4EF - Data space Integration}
    \label{fig:dataspace-integration}
\end{figure}

AI4EF connects to the European Energy Data Space both as a data source for model training in the Training Playground and through integration with the Enershare Data Space Marketplace. This enables ingestion of real-time data from sources like smart meters and building systems, enhancing model accuracy and customization. The link to Enershare Data Space also aligns AI4EF with European data-sharing standards, supporting secure, compliant data access for stakeholders in energy efficiency projects. Figure \ref{fig:dataspace-integration} illustrates the data space integration framework of AI4EF, detailing connections with the Enershare Data Space Marketplace.

\begin{figure}[!h]
    \centering
    \includegraphics[scale=0.3]{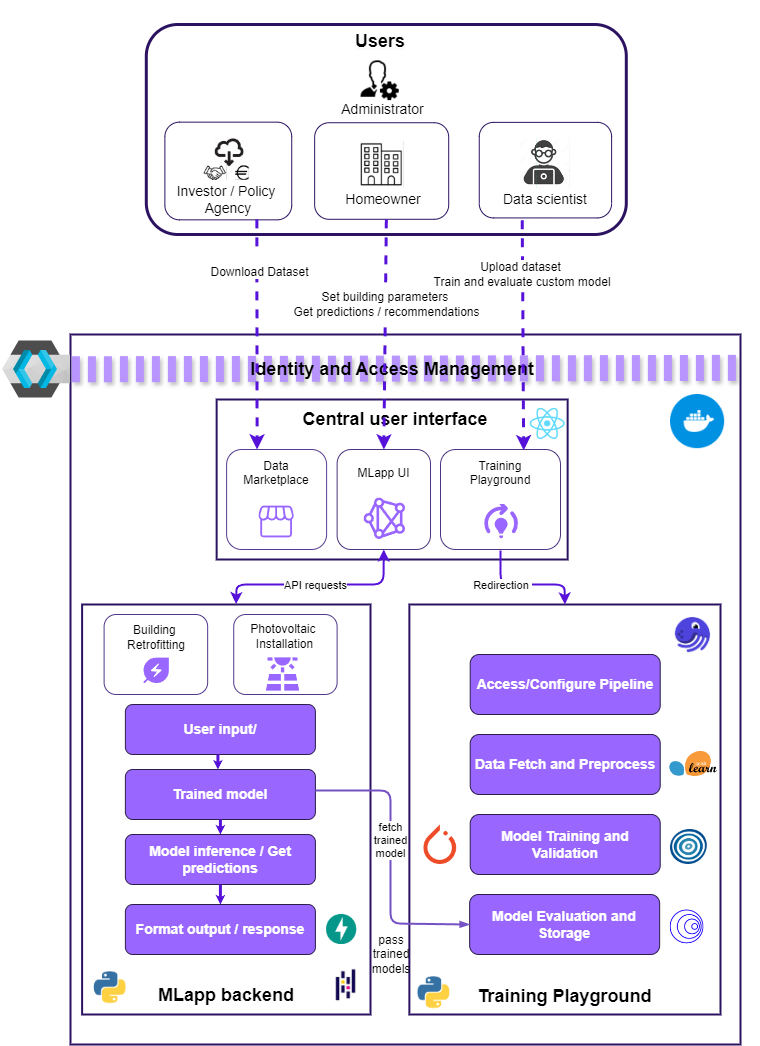}
    \caption{AI4EF - Software Architecture Diagram}
    \label{fig:architecture}
\end{figure}

\subsection{Software functionalities}


\subsection{Interface Features}
AI4EF can be accessed and be navigated through its main UI, which serves as a centralized dashboard for all sub-services of the tool.  

\begin{figure}[!h]
    \centering
    \begin{subfigure}{0.46\textwidth} 
        \centering
        \includegraphics[width=\linewidth,height=6cm,keepaspectratio]{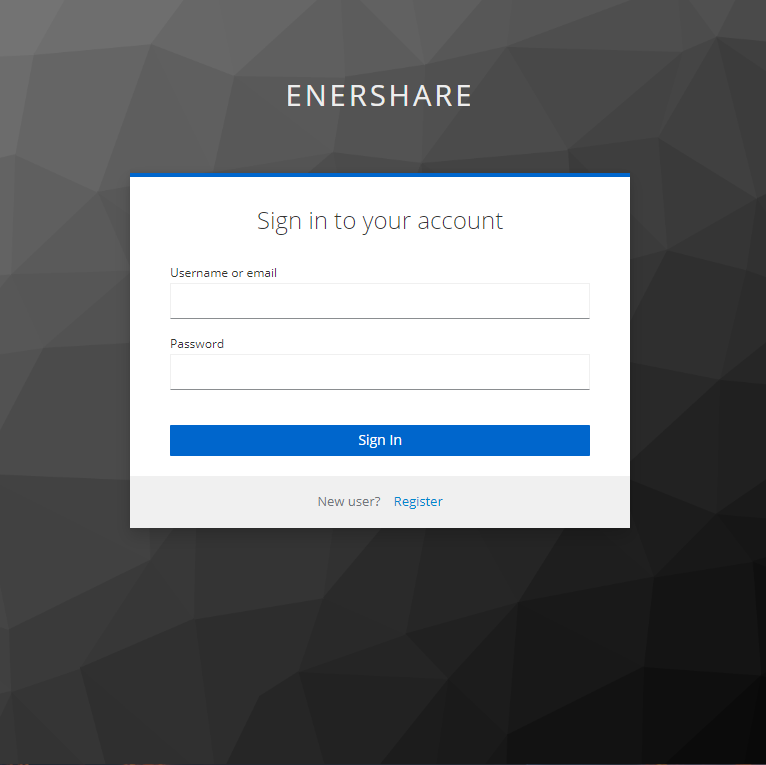}
        \caption{Keycloak sign-in page}
        \label{fig:keycloak}
    \end{subfigure}\hfill
    \begin{subfigure}{0.46\textwidth} 
        \centering
        \includegraphics[width=\linewidth,height=6cm,keepaspectratio]{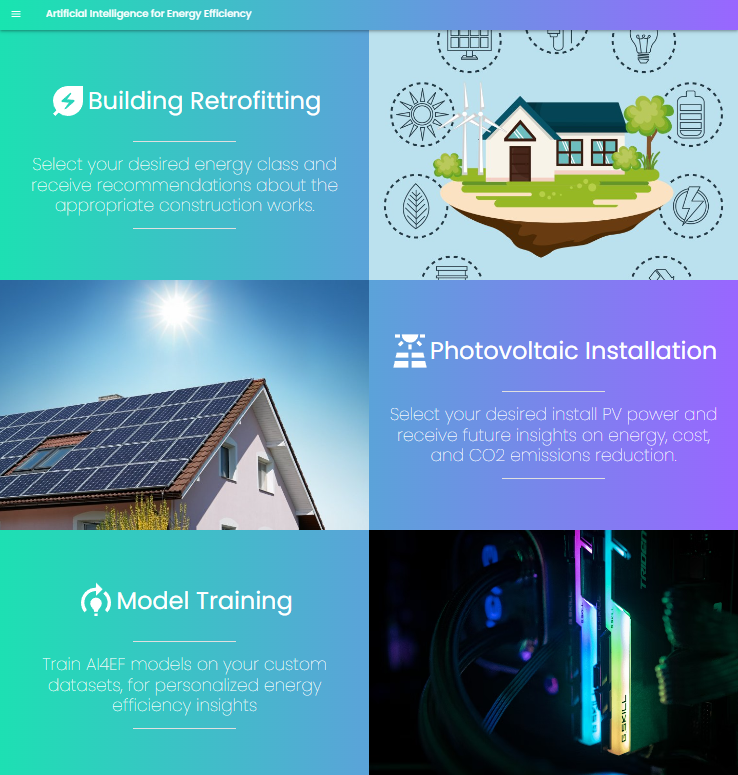}
        \caption{AI4EF Homepage}
        \label{fig:homepage}
    \end{subfigure}
    
    \vspace{0.2cm} 

    \begin{subfigure}{0.46\textwidth} 
        \centering
        \includegraphics[width=\linewidth,height=6cm,keepaspectratio]{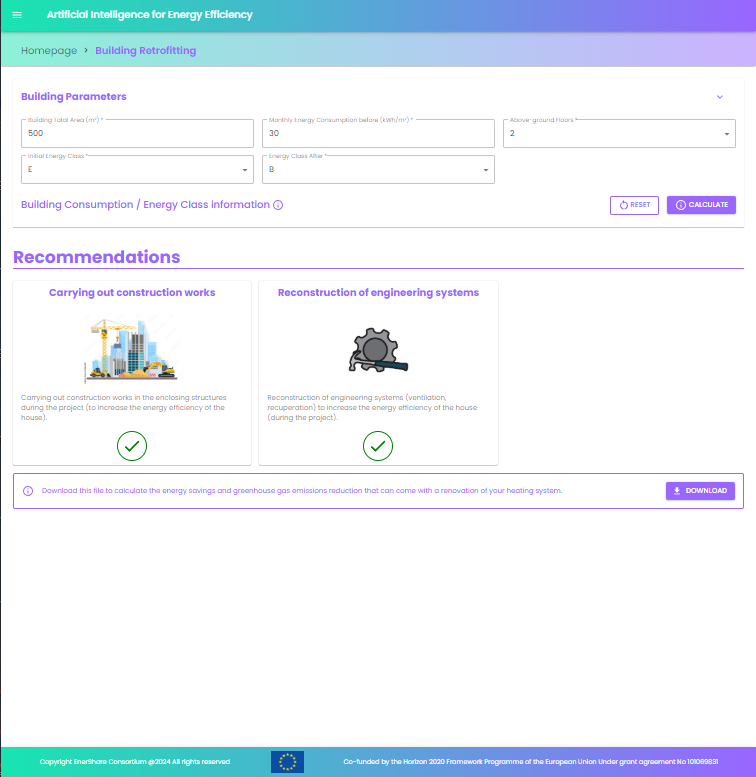}
        \caption{AI4EF - Building Retrofitting Example}
        \label{fig:ai4ef-building-retrofitting-full}
    \end{subfigure}\hfill
    \begin{subfigure}{0.46\textwidth} 
        \centering
        \includegraphics[width=\linewidth,height=6cm,keepaspectratio]{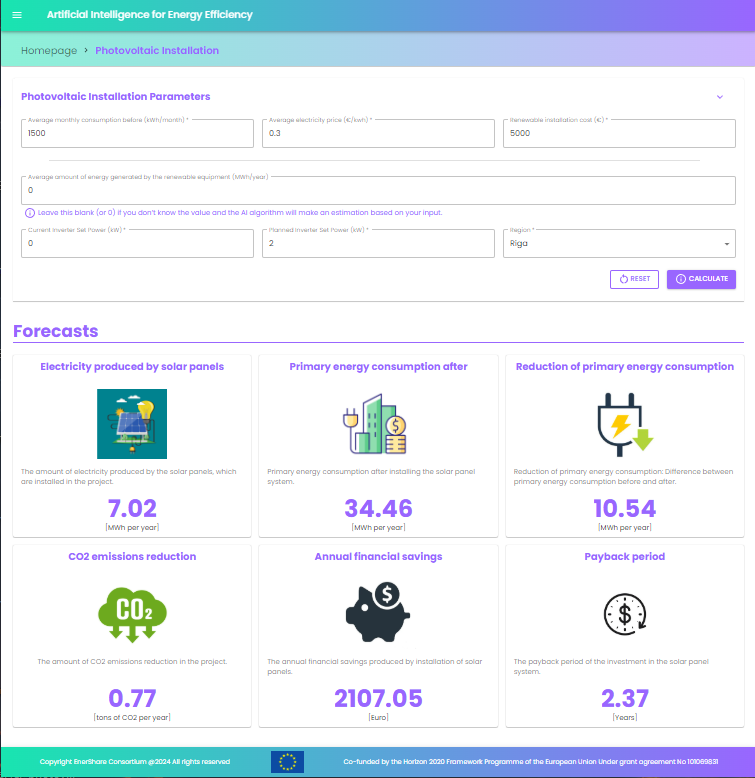}
        \caption{AI4EF - Photovoltaic Installation Example}
        \label{fig:ai4ef-photovoltaic-installation-full}
    \end{subfigure}

    \caption{AI4EF - MLApp Dashboard}
\end{figure}

\paragraph{\textbf{Homepage:}}
Figure \ref{fig:homepage} shows the homepage displayed after signing in. Here, an overview of the primary tools is presented in a visually organized layout, with clear headings that introduce the key services. 

\paragraph{\textbf{MLApp - Building Retrofitting:}} 
The Building Retrofitting sub-service provides tools to analyse the potential for building retrofits and energy upgrades. 


\paragraph{\textbf{MLApp - Photovoltaic Installation:}}

The Photovoltaic Installation sub-service evaluates the economic and environmental impact of PV installations on energy savings and emissions reductions. 


\begin{table}[ht!]
\centering
\caption{MLApp Input/Output Parameters}
\label{tab:mlapp-feature-targets}
    \begin{tabular}{>{\centering\arraybackslash}p{3cm} >{\centering\arraybackslash}p{3.5cm} >{\centering\arraybackslash}p{3.5cm} >{\centering\arraybackslash}p{4cm} } \toprule
    \multicolumn{2}{c}{\textbf{Building Retrofitting}} & \multicolumn{2}{c}{\textbf{Photovoltaic Installation}} \\ \midrule
    \textbf{Inputs} & \textbf{Targets} & \textbf{Inputs} & \textbf{Targets} \\ \midrule 
    \rowcolor{gray!10} 
    \makecell[c]{Building \\ Total Area} & \makecell[c]{Carrying out \\ construction works} & \makecell[c]{Average electricity \\ price for 1 kW} & \makecell[c]{Electricity produced \\ by solar panels} \\[0.6cm] 
    \makecell[c]{Above-ground \\ floors} & \makecell[c]{Reconstruction of \\ engineering systems} & \makecell[c]{Average monthly \\ consumption before} & \makecell[c]{Primary energy \\ consumption after} \\[0.6cm]
    \rowcolor{gray!10}
    \makecell[c]{Energy \\ consumption \\ before} & Heat installation & \makecell[c]{Installation costs of \\ renewable production \\ equipment} & \makecell[c]{Reduction of primary \\ energy consumption} \\[0.6cm]
    Initial energy class & \makecell[c]{Water heating system} & Current inverter set power & CO2 emissions reduction \\[0.6cm]
    \rowcolor{gray!10}
    Energy class after renovation &  & Inverter power in project & \makecell[c]{Expected annual \\ self consumption} \\ 
     &  & \makecell[c]{Average amount of \\ energy generated by \\ the equipment} & Annual financial savings \\[0.6cm] 
    \rowcolor{gray!10} 
     &  & Region & Payback period \\[0.6cm] 
     \bottomrule
    \end{tabular}

\end{table}

\subsubsection{Training Playground service}
\begin{figure}[!h]
    \centering
    \includegraphics[width=0.8\linewidth]{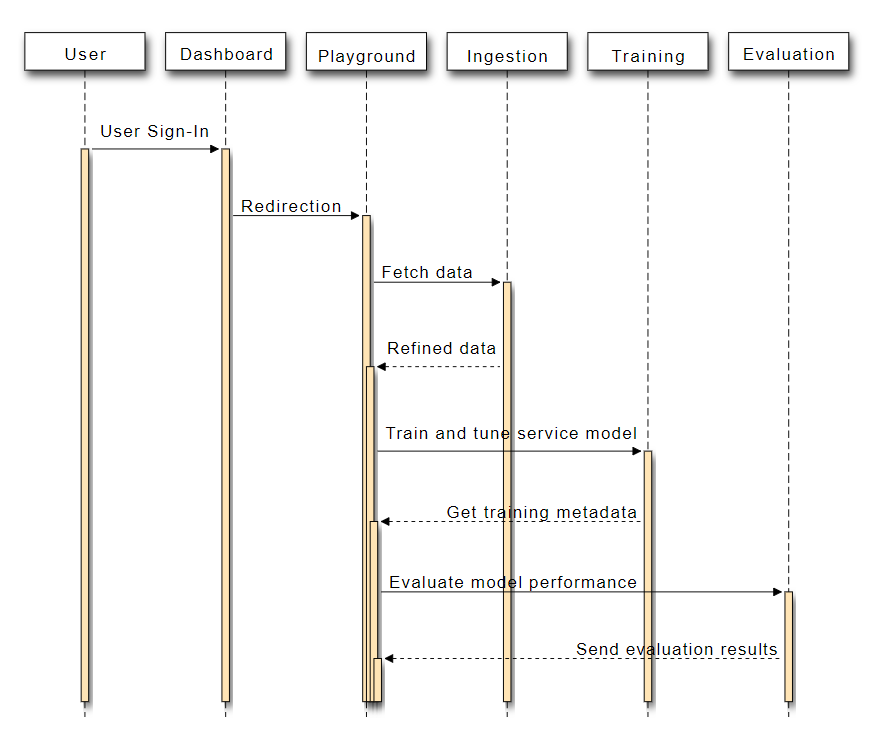}
    \caption{AI4EF - Training Playground Sequence Diagram}
    \label{fig:playground-seqdiag}
\end{figure}

The \textbf{Training Playground} is an ML environment for building and optimizing ML models that predict energy outcomes and support energy efficiency projects. Data scientists can execute complete ML workflows, from data acquisition and preprocessing to hyperparameter tuning and evaluation, via Dagster. The pipeline comprises 3 consecutive steps, as illustrated in figure \ref{fig:playground-seqdiag}: the \textit{Ingestion}, the \textit{Training} and the \textit{Evaluation}.

\paragraph{\textbf{Ingestion}}
This step is separated into two distinct categories: the helper functions and the main assets. Starting from the helper functions, these include methods for validating file paths, URLs, and database connection strings, as well as for retrieving data from various sources like databases, local files, or APIs. The main assets handle data extraction, cleaning, and scaling. They support loading data, managing categorical and continuous scaling, handling missing values, and creating train-test splits. 

\paragraph{\textbf{Training}}
This step sets up an automated pipeline to train and optimize ML models using Optuna for hyperparameter tuning, PyTorch Lightning for model management, and Dagster for orchestration. It experiments with a Multi-Layer Perceptron (MLP) architecture, running multiple trials to test different configurations (e.g., layers, learning rate) to find the best model. Early stopping and pruning halt unpromising trials, and the top model and preprocessing settings are saved for deployment. Dagster’s asset management ensures modular, scalable, and reusable components for efficient experimentation.

\paragraph{\textbf{Evaluation}}
This step sets up an evaluation pipeline to test model performance, generate metrics, and visualize HPO results. It computes different evaluation metrics depending on the sub-service: regression or classification. Visual summaries of the HPO process show parameter importance and performance trends. All metrics and plots are stored in a structured format, providing a clear assessment of model accuracy and tuning success, facilitating validation and deployment.

\subsubsection{Enershare Data Space Marketplace}

The \textbf{Enershare Data Space Marketplace} is a repository for data assets accessible directly from the AI4EF platform. Users can import and export data artifacts for enhanced analysis and collaboration, supporting data-driven decision-making across energy efficiency projects.

\subsubsection{User roles}

The AI4EF platform is designed to support various stakeholders with distinct roles. Government representatives use it for policy-making and regulatory compliance, while building managers focus on energy optimization and retrofitting. Energy consultants provide expertise on efficiency measures, and data scientists analyse trends and build predictive models. Each role has tailored functionalities to collaborate and access relevant data, enhancing the platform’s overall effectiveness. This role-based approach, as shown in Figure \ref{fig:architecture}, ensures a comprehensive ecosystem for sustainable energy planning and investment.

\section{Illustrative examples}


\subsection{Overview of the MLApp Dashboard}
The MLApp Dashboard features two individual sub-services: \textit{Building Retrofitting} and \textit{Photovoltaic Installation}. Both services share a similar UI, allowing for easy navigation and data input, while the specific input and output fields vary according to each service’s focus. The specific input values for each example are provided in Table \ref{tab:mlappp-illustration-example-inputs} of the appendix for reference.

In the \textbf{Building Retrofitting} sub-service, users begin by filling out the Building Parameters form, and then click the \textit{CALCULATE} button to send their information to the backend ML model. The results and recommendations are illustrated in Figure \ref{fig:ai4ef-building-retrofitting-full}. Users can also download an Excel file detailing potential energy savings related to suggested upgrades and use the \textit{RESET} button to clear the form as needed.

Similarly, the \textbf{Photovoltaic Installation} sub-service prompts users to input relevant details in the respective form. Users may leave the energy generation field blank for automatic predictions based on other inputs. 
The user flow can be easily inferred from Figure \ref{fig:ai4ef-photovoltaic-installation-full}.



    

    

\subsection{Overview of the Training Playground}
With the Training Playground service, users access the Training Playground from the AI4EF Dashboard's sidebar. They will then be redirected to the \textit{Overview - Jobs} page, where they can navigate to the Dagster UI.
They can choose to run each step of the pipeline independently, 
or execute the entire pipeline in one go, as presented in the overview. In Figure \ref{fig:playground-job_ml_pipeline}, we illustrate the individual pipelines and their interconnectivity.
After that, users configure the pipeline parameters through the built-in Launchpad menu. In Table \ref{tab:dagster-configuration} of the appendix, we have compiled the parameters that users can configure through the Launchpad, each of which is essential for different steps of the pipeline. 

\begin{figure}[!h]
    \centering
    \begin{subfigure}{0.46\textwidth}
        \centering
        \includegraphics[width=\linewidth]{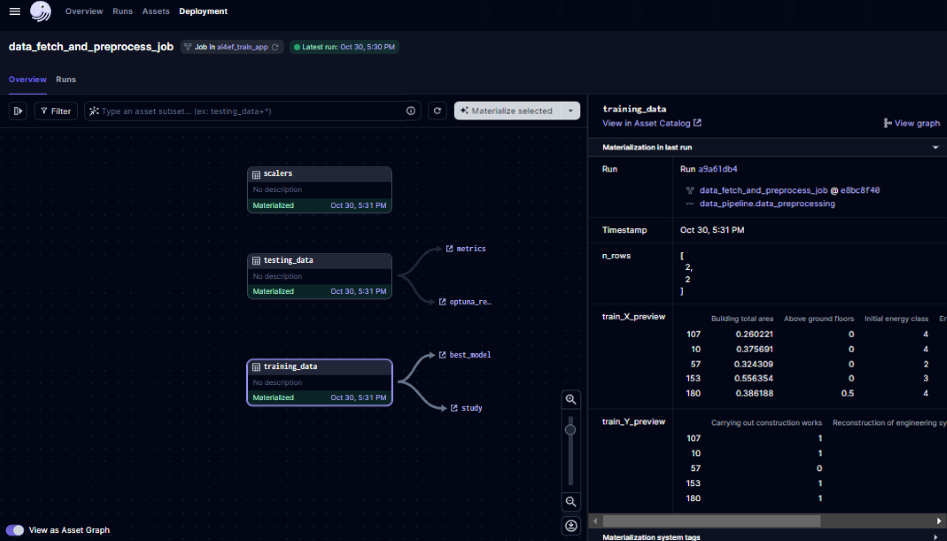}
        \caption{AI4EF - Training Playground - Data Ingestion step}
        \label{fig:playground-ingestion}
    \end{subfigure}\hfill
    \begin{subfigure}{0.46\textwidth} 
        \centering
        \includegraphics[width=\linewidth]{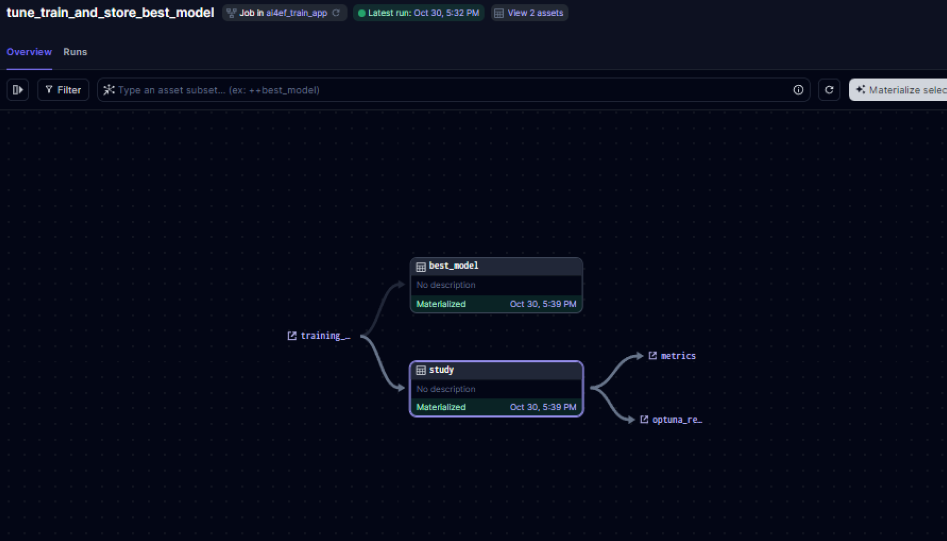}
        \caption{AI4EF - Training Playground - Training Pipeline Overview}
        \label{fig:playground-train-overview} 
    \end{subfigure}
    
    \vspace{0.5cm} 

    \begin{subfigure}{\textwidth} 
        \centering
        \includegraphics[width=\linewidth]{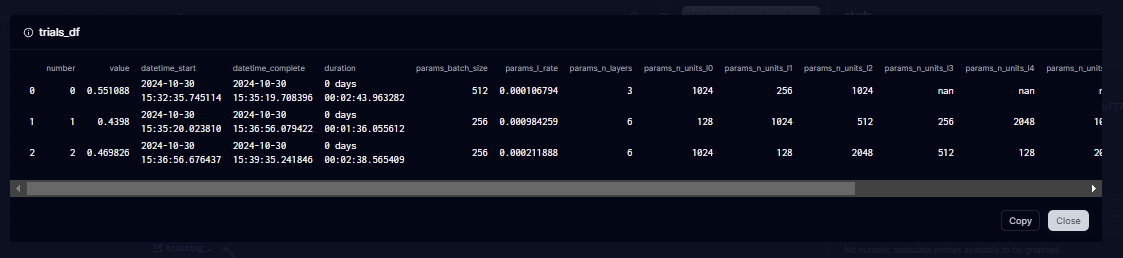}
        \caption{AI4EF - Training Playground - Training Pipeline Trials}
        \label{fig:playground-train-trials} 
    \end{subfigure}
    
    \caption{AI4EF - Training Playground - Ingestion and Training Pipelines}
    \label{fig:playground-train} 
\end{figure}

\paragraph{\textbf{Data pipeline (Ingestion)}}
In the data pipeline step, AI4EF has produced three Dagster objects: train-data, test-data, and scalers and several metadata as depicted in Figure \ref{fig:playground-ingestion}. 


\paragraph{\textbf{Training Pipeline (Training)}}
During model tuning, AI4EF produces two key outputs: the best model from hyperparameter tuning and a detailed study of the tuning process. As shown in Figure \ref{fig:playground-train}, AI4EF provides a Pandas \cite{pandas} dataframe documenting each tuning trial with details on trial number, start and end times, duration, and hyperparameters like batch size and learning rate. 

\begin{figure}[!h]
    \centering
    \begin{subfigure}{0.46\textwidth} 
        \centering
        \includegraphics[width=\linewidth, height=0.8\textwidth]{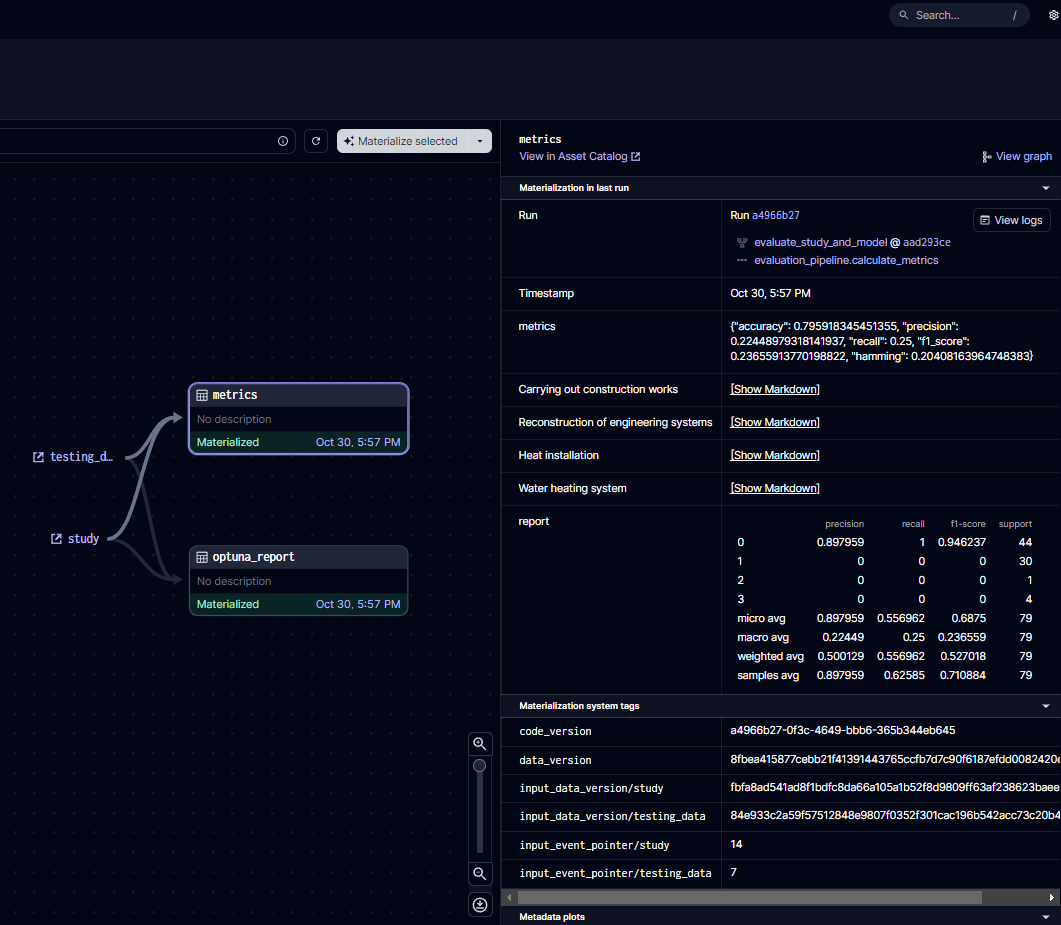}
        \caption{Evaluation Metrics Overview}
        \label{fig:playground-evaluation-evaluation_metrics}
    \end{subfigure}
    \hspace{0.001\textwidth} 
    \begin{subfigure}{0.46\textwidth} 
        \centering
        \includegraphics[width=\linewidth, height=0.8\textwidth]{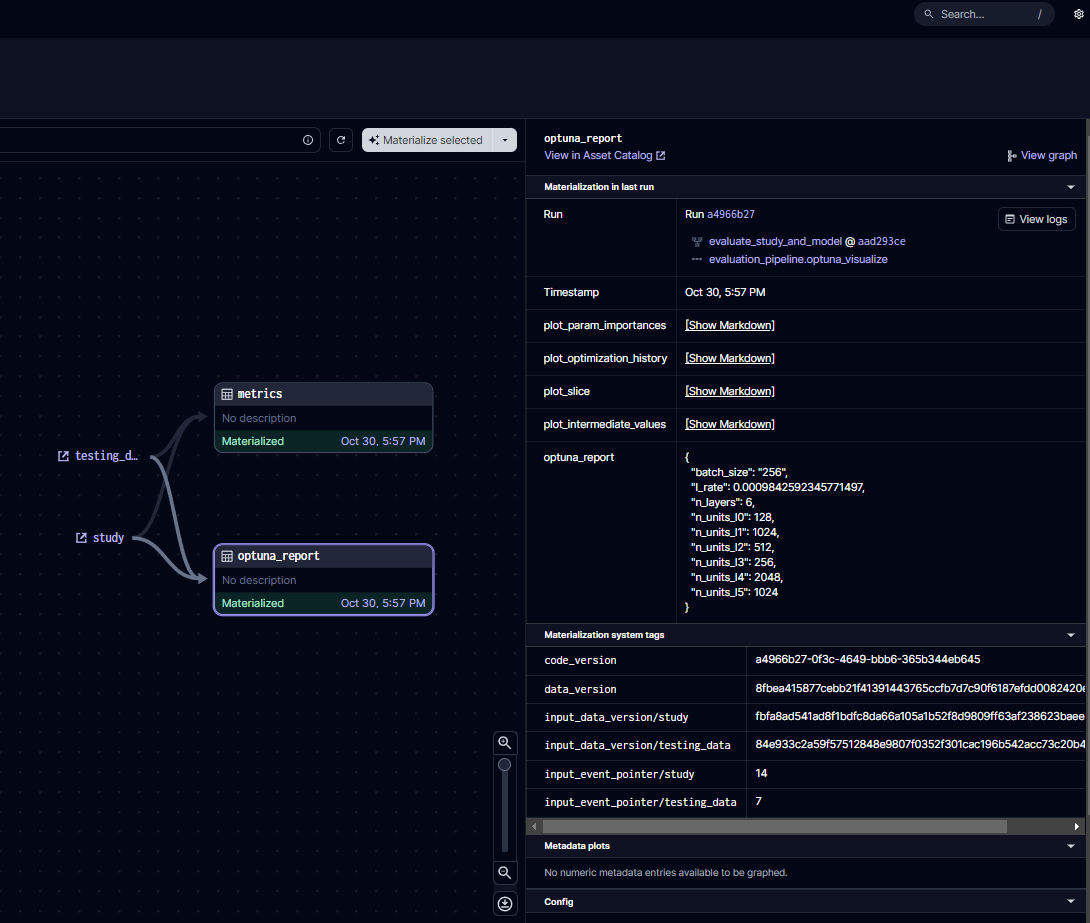}
        \caption{Optuna Optimization Results}
        \label{fig:playground-evaluation-optuna_results}
    \end{subfigure}

    \vspace{0.4cm} 

    \begin{subfigure}{0.46\textwidth} 
        \centering
        \includegraphics[width=\linewidth, height=0.8\textwidth]{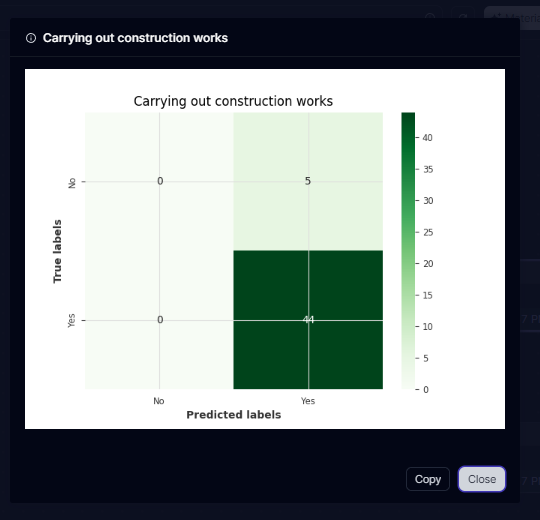}
        \caption{Confusion Matrix for Classifier Performance in 'Carrying out construction works' target}
        \label{fig:playground-evaluation-confusion_matrix}
    \end{subfigure}
    \hspace{0.001\textwidth} 
    \begin{subfigure}{0.46\textwidth} 
        \centering
        \includegraphics[width=\linewidth, height=0.8\textwidth]{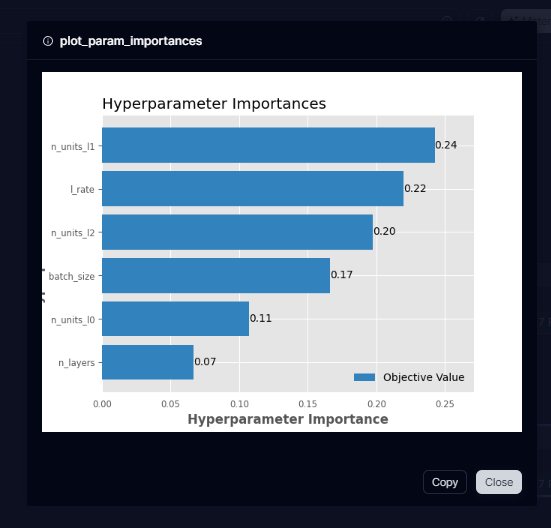}
        \caption{Parameter Importance during model training and tuning}
        \label{fig:playground-evaluation-param_importance}
    \end{subfigure}

    \caption{AI4EF - Evaluation pipeline}
    \label{fig:playground-evaluation}
\end{figure}

\paragraph{\textbf{Evaluation pipeline (Evaluation):}}
In the final step, AI4EF generates two key objects for model evaluation. The \textit{metrics} object, adapts based on the model type: for classifiers, it includes metrics such as accuracy, precision, recall, and confusion matrices (see Figures \ref{fig:playground-evaluation-evaluation_metrics} and \ref{fig:playground-evaluation-confusion_matrix}).
The second object details hyperparameter tuning with Optuna, including parameter importance, optimization history, and optimal values, as shown in Figures \ref{fig:playground-evaluation-optuna_results} and \ref{fig:playground-evaluation-param_importance}. 

\section{Impact}
The AI4EF web application is designed to significantly impact energy efficiency and retrofitting efforts in the building sector. By providing a data-driven, user-friendly tool for decision-making, AI4EF facilitates the identification of energy-saving opportunities, ultimately leading to reduced energy consumption and lower operational costs for building managers and owners. 

The platform supports government representatives in formulating evidence-based policies and regulations aimed at promoting sustainable energy practices. Furthermore, energy consultants can leverage AI4EF to enhance their advisory services, providing clients with tailored insights and actionable strategies for implementing energy efficiency measures.

By integrating advanced ML techniques and predictive analytics, AI4EF empowers users to make informed decisions regarding investments in renewable energy sources and retrofitting projects, thereby fostering the transition toward more sustainable building practices. The modular design of the platform allows for continuous updates and improvements, ensuring that users benefit from the latest technological advancements and data insights.

AI4EF has demonstrated significant real-world impact, particularly within the EU's Horizon Europe research project, Enershare, where it is designed to optimize energy efficiency and support data-driven decision-making for building retrofitting in Latvia. Reviewed and used by actual users to meet their objectives, and it was developed using real data based on the requirements of LEIF. This guarantees that the technology satisfies actual needs and provides a workable solution for building energy efficiency problems. The tool's integration into Enershare highlights its potential for broader application. AI4EF can be applied across diverse European countries, adapting to local energy landscapes and helping drive the transition to an energy-efficient built environment.


\section{Conclusions and Future Work}

In this paper, we introduced AI4EF, a software application designed
to support AI-driven decision-making in the energy efficiency sector. It offers various energy efficiency goals, such as investment planning and photovoltaic installation insights, while also offering flexibility for those needing custom ML models tailored their unique data. Using MLOps best practices and advanced algorithms, AI4EF simplifies data preprocessing, model training, evaluation and deployment, making it an accessible tool for data scientists, energy consultants, and decision-makers alike.

Looking ahead, AI4EF's development will focus on enhancing the flexibility and scalability of its services. A key area for future work is the integration of additional data sources from multiple countries, broadening the tool's applicability across Europe. We also aim to refine the user interface to create more intuitive interactions. Furthermore, we plan to boost model performance by incorporating advanced ML techniques, such as transfer learning, to tackle user cases where data are scarce or exotic. Additionally, we plan to enhance model explainability, making it easier for users to understand the factors influencing predictions and providing greater transparency. Support for model version control will also be implemented, for better user interactions across the platform’s core services. Through these advancements, AI4EF aims to play a central role in driving the transition to a sustainable, energy-efficient future.

\section*{Acknowledgements} 
\label{acknowledgments}
This work has been funded by the European Union’s Horizon Europe Research and Innovation programme under the Enershare project, grant agreement No. 101069831. The sole responsibility for the content of this paper lies with the authors; the paper does not necessarily reflect the opinion of the European Commission.\\

\newpage

\bibliographystyle{unsrt} 
\bibliography{references}


\pagebreak




\begin{appendices}
\section*{Illustrative Example inputs}

\begin{table}[ht!]
    \centering
    \caption{Building Retrofitting and Photovoltaic Installation Example Inputs}
    \label{tab:mlappp-illustration-example-inputs}
    \begin{tabular}{>{\centering\arraybackslash}p{3.5cm} >{\centering\arraybackslash}p{3.5cm} | >{\centering\arraybackslash}p{3.5cm} >{\centering\arraybackslash}p{3.5cm} }
        \toprule
        \multicolumn{2}{c|}{\textbf{Building Retrofitting}} & \multicolumn{2}{c}{\textbf{Photovoltaic Installation}} \\ \midrule
        \textbf{Parameter} & \textbf{Value} & \textbf{Parameter} & \textbf{Value} \\ \midrule
        \rowcolor{gray!10} 
        Building Total Area & 500 & Average Monthly Consumption Before & 1500 \\[0.6cm]
        Above-ground Floors & 2 & Average Electricity Price & 0.3 \\[0.6cm]
        \rowcolor{gray!10}
        Energy Consumption Before & 30 & Installation Costs of Renewable Equipment & 5000 \\[0.6cm]
        Initial Energy Class & E & Average Amount of Energy Generated by Renewable Equipment & - \\[0.6cm]
        \rowcolor{gray!10}
        Energy Class After & B & Current Inverter Set Power & 0 \\[0.6cm]
         &  & Planned Inverter Set Power & 2 \\[0.6cm]
        \rowcolor{gray!10}
         &  & Region & Riga \\ 
        \bottomrule
    \end{tabular}
\end{table}

\vspace{-2cm}

\begin{table}[ht!]
    \centering
    \small
    \caption{Configuration Parameters for Dagster Dashboard}
    \label{tab:dagster-configuration} 
    \resizebox{0.8\textwidth}{0.7\textwidth}{
        \begin{tabular}{>{\centering\arraybackslash}p{0.2\linewidth} >{\centering\arraybackslash}p{0.3\linewidth} >{\centering\arraybackslash}p{0.3\linewidth} >{\centering\arraybackslash}p{0.2\linewidth}}
            \toprule
            \textbf{Parameter} & \textbf{Value} & \textbf{Description} & \textbf{Pipeline Step}\\ \midrule
            \rowcolor{gray!10} 
            activation & \makecell[c]{ReLU} & \makecell[c]{Activation function \\used in the model} & Training \\
            authorization & \makecell[c]{APIKEY-xxxxx...xxxx} & \makecell[c]{API key \\for authorization} & Ingestion \\
            \rowcolor{gray!10}
            batch\_size & \makecell[c]{256, 512, 1024} & \makecell[c]{Batch sizes \\ for training the model} & Training \\
            consumer\_agent\_id & \makecell[c]{urn:ids:enershare:\\connectors:www:xxx} & \makecell[c]{Identifier for the \\ consumer agent} & Ingestion \\
            \rowcolor{gray!10}
            feature\_cols & \makecell[c]{Building total area,\\Above ground floors, \\Initial energy class,\\Energy consumption before,\\Energy class after} & \makecell[c]{\\ \\Feature columns \\used in the model} & Training \\
            input\_filepath & \makecell[c]{https://\textless baseurl\textgreater/\\\textless data-app-path\textgreater\\/openapi/\textless version\textgreater/\\\textless endpoint\textgreater} & \makecell[c]{URL for the input \\data file} & Ingestion \\
            \rowcolor{gray!10}
            l\_rate & \makecell[c]{0.0001, 0.001} & \makecell[c]{Learning rates \\for the model} & Training \\
            layer\_sizes & \makecell[c]{128, 256, 512, \\1024, 2048} & \makecell[c]{\\ Range of suggested \\ model layer sizes} & Training \\
            \rowcolor{gray!10}
            max\_epochs & \makecell[c]{10} & \makecell[c]{Maximum number \\of training epochs} & Training \\
            mlClass & \makecell[c]{Classifier} & \makecell[c]{ML \\class used} & Training \\
            \rowcolor{gray!10} 
            \makecell[c]{ml\_path} & \makecell[c]{/leif\_app/shared\_storage/\\models-scalers/\\best\_MLPClassifier.ckpt } & \makecell[c]{Path to saved \\model checkpoint} & Evaluation \\
            n\_layers & \makecell[c]{2, 6} & \makecell[c]{Number of layers \\ in the model} & Training \\
            \rowcolor{gray!10}
            n\_trials & \makecell[c]{3} & \makecell[c]{Number of trials for \\hyperparameter tuning} & Training \\
            num\_workers & \makecell[c]{2} & \makecell[c]{Number of workers \\for data loading} & Ingestion \\
            \rowcolor{gray!10}
            optimizer\_name & \makecell[c]{Adam} & \makecell[c]{Optimizer used for model training} & Training \\
            \makecell[c]{optuna\_viz} & \makecell[c]{/leif\_app/shared\_storage/\\optuna\_viz/classifier/} & \makecell[c]{Directory for Optuna \\visualization files} & Evaluation \\
            \rowcolor{gray!10}
            provider\_agent\_id & \makecell[c]{urn:ids:enershare:\\connectors:yyy:zzz} & \makecell[c]{Identifier for the provider agent} & Ingestion \\
            scalers\_path & \makecell[c]{/leif\_app/shared\_storage/\\models-scalers/\\MLPClassifier\_scalers.pkl} & \makecell[c]{Path to the scaler file \\ for normalization} & Training \\
            \rowcolor{gray!10}
            seed & \makecell[c]{42} & \makecell[c]{Random seed for \\reproducibility} & Training \\
            target\_cols & \makecell[c]{\footnotesize Carrying out \\construction works,\\ \footnotesize Reconstruction of \\engineering systems,\\\footnotesize Heat installation,\\ Water heating system} & \makecell[c]{List of target columns \\for predictions} & Training \\
            \bottomrule
        \end{tabular}
    }
\end{table}

\section*{Supplementary Figures}

\begin{figure}[!h]
    \centering
    \includegraphics[width=\linewidth, height=0.8\textwidth]{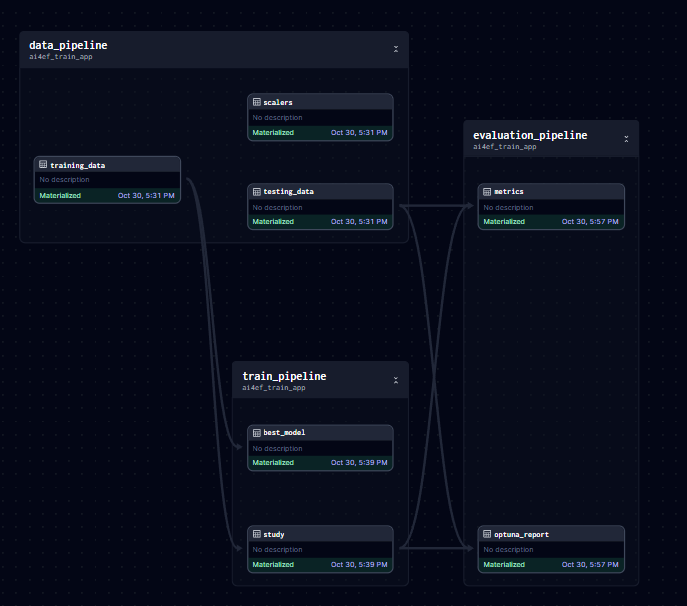}
    \caption{AI4EF - Training Playground - ML Pipeline}
    \label{fig:playground-job_ml_pipeline}
\end{figure}

\begin{figure}[!h]
        \centering
        \includegraphics[width=\linewidth]{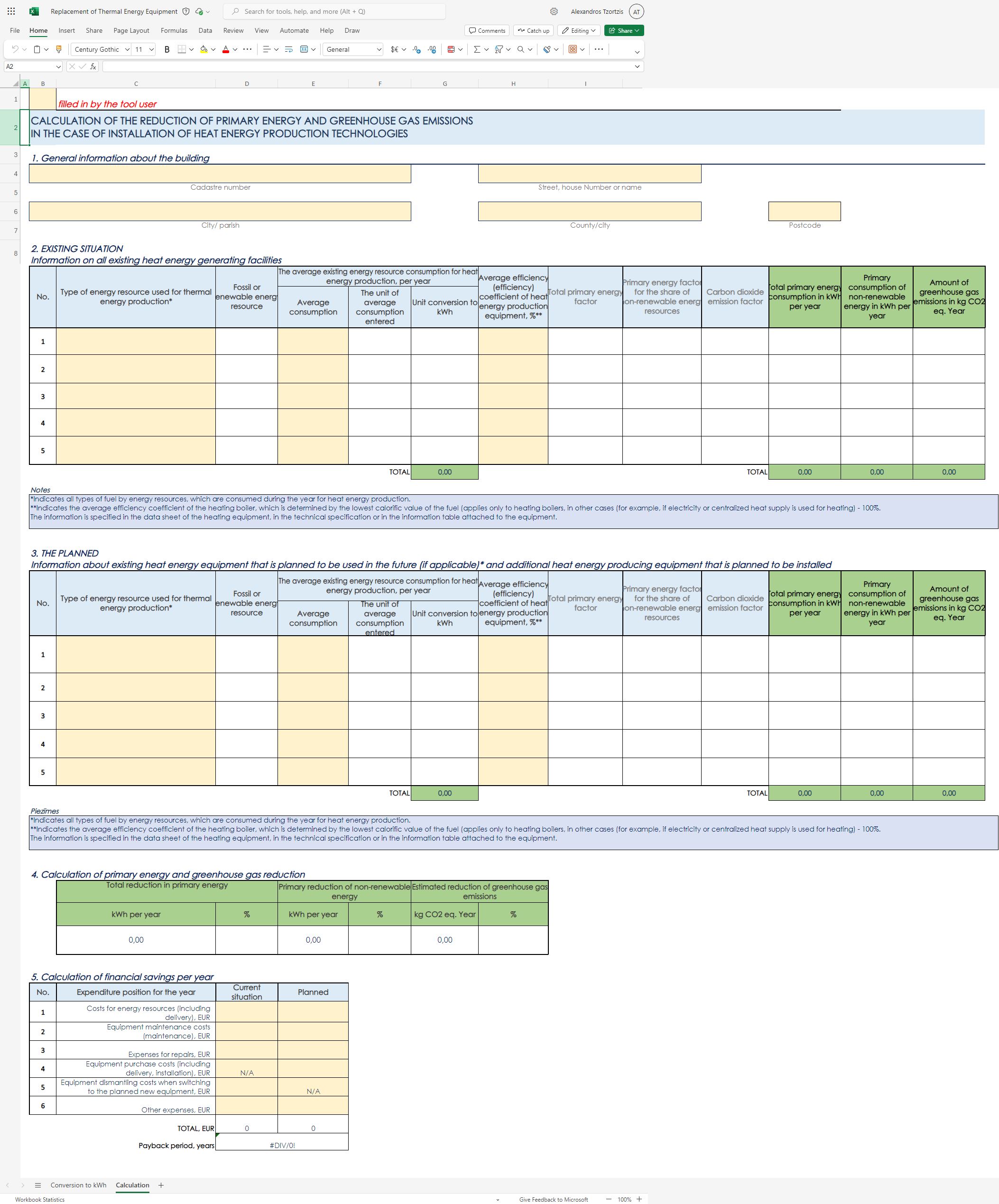}
        \caption{AI4EF -  Excel sheet used to calculate energy savings and greenhouse gas emissions reduction that can come with a renovation of your heating system.}
        \label{fig:thermal-energy-excel}
\end{figure}
\end{appendices}




\end{document}